\documentclass[conference]{IEEEtran}
\IEEEoverridecommandlockouts
\usepackage{amsmath,amssymb,amsfonts}
\usepackage{graphicx}
\usepackage[caption=false, font=footnotesize]{subfig}
\usepackage{textcomp}
\usepackage{xcolor}
\usepackage{multirow}
\usepackage{booktabs}
\usepackage{bm}
\usepackage[
backend=biber,
style=ieee
]{biblatex}
\def\BibTeX{{\rm B\kern-.05em{\sc i\kern-.025em b}\kern-.08em
    T\kern-.1667em\lower.7ex\hbox{E}\kern-.125emX}}
\begin{document}

\renewcommand*{\bibfont}{\normalfont\small}

\title{Improving 3D Labeling in Self-Driving by Inferring Vehicle Information using Vision Language Models\thanks{Accepted at the IEEE Intelligent Vehicles Symposium (IV) 2026.}}

\author{
\IEEEauthorblockN{Steven Chen}
\IEEEauthorblockA{\textit{Aurora Innovation, Inc.} \\
Mountain View, CA \\
schen@aurora.tech}
\and
\IEEEauthorblockN{Shivesh Khaitan}
\IEEEauthorblockA{\textit{Aurora Innovation, Inc.} \\
San Francisco, CA \\
skhaitan@aurora.tech}
\and
\IEEEauthorblockN{Nemanja Djuric}
\IEEEauthorblockA{\textit{Aurora Innovation, Inc.} \\
Pittsburgh, PA \\
ndjuric@aurora.tech}
}

\maketitle

\begin{abstract}
We present an approach to improve 3D vehicle labeling in self-driving applications through zero-shot inference of vehicle information, leveraging Vehicle Make and Model Recognition (VMMR) methods. 
The proposed approach utilizes a Vision Language Model (VLM) to both infer a vehicle’s make, model, and generation from image crops, and output accurate 3D bounding box dimensions to seed manual labeling.
We evaluate the impact of iterative prompt engineering and the choice of different VLMs on both vehicle bounding box inference and make/model/generation recognition. 
When compared to strong baselines, the proposed approach not only shows high accuracy but also excels in mitigating specific failure modes where VLMs provide better dimensions than initial lidar-aided human-annotated labels (e.g., in cases of significant vehicle occlusion). 
Experiments on both public and proprietary data strongly suggest that our conclusions are generalizable across different labelers and datasets. 
The results demonstrate that integrating VLMs into the labeling process can reduce manual labeling time while increasing label quality.
\end{abstract}


\section{Introduction}

Autonomous vehicle (AV) technology is increasingly present on public roads \cite{favaro2025determining}, powered by machine learning (ML) models that are integral to systems such as perception \cite{sumalatha2024autonomous, fawole2024recent} and motion planning \cite{ganesan2024comprehensive}.
Similarly to ML models in other industries, having large amounts of high-quality data is critical for model performance and generalizability \cite{sun2020waymo}.
Traditionally, researchers relied on manual annotation, which involves drawing accurate 3D bounding boxes by hand around vehicles, pedestrians,  and other traffic actors in a scene \cite{Geiger2021KITTI}.
However, this manual process is tedious and challenging, as AV data is large, complex, and multi-modal (including sensors such as cameras, lidars, and radars) \cite{huang2024detecting,domhof2021joint}.
Therefore, auto-labeling approaches that automatically infer and seed bounding boxes before manual labeling are attractive alternatives to speed up and improve the labeling process \cite{qi2021offboard,houston2021one}.
We explore this topic and propose a novel method to improve auto-labeling, leveraging recent advances in foundation models.

A recent development that has impacted the AV industry is the rise of foundation models, methods that are largely self-supervised on vast unlabeled datasets, shown to lead to improved performance on a number of tasks \cite{zhou2024comprehensive}. 
A special kind of foundation models are Vision Language Models (VLMs) \cite{chen2024wayve}, which can take multiple modalities as inputs (e.g., camera images, lidar sweeps, text) and are thus particularly applicable to self-driving due to the vast quantities of multi-sensor data that are generated in the field.
For example, there exist initial attempts to apply VLMs to end-to-end driving, aiming to explain decisions of black-box self-driving models or even to try directly outputting vehicle controls \cite{marcu2024lingoqa, Wayve_2023}.
Beyond these onboard applications, their strong performance in computer vision tasks \cite{kirillov2023segment, ravi2025sam} makes VLMs promising for offboard 3D auto-labeling as well, with initial work exploring two-stage processing of VLM 2D outputs \cite{openannotate3d,openbox}.

We propose a method to significantly improve auto-labeling speed and quality for self-driving by directly inferring vehicle 3D dimensions, especially relevant in critical cases where sensor data is sparse.
Moreover, the proposed approach connects the research fields of vehicle make and model recognition (VMMR) and auto-labeling in a novel way, using make, model, and generation information as a strong prior for vehicle extents.
Our main contributions are summarized below:
\begin{itemize}
    \item We propose a VLM-based method that, within a single pass, performs two coupled tasks: inferring a vehicle's make, model, and generation, and outputting accurate 3D vehicle dimensions to seed label bounding boxes;
    \item We demonstrate an iterative prompt engineering process that improves accuracy (e.g., by adding specific reasoning around vehicle generation and exterior modifications);
    \item We evaluate the proposed method using multiple state-of-the-art VLMs and against strong baselines, and validate our results on both proprietary and open-sourced data, showing large performance improvements;
    \item We identify and analyze failure cases, showing that our approach can provide more accurate dimensions than initial lidar-aided human-annotated labels in challenging scenarios, such as in the case of occluded vehicles.
\end{itemize}

\section{Related Work}
While recent large-scale, open-sourced datasets from industry have become the backbone of AV research \cite{sun2020waymo,wilson2023argoverse}, the 3D labeling process remains a manual, costly, and complex bottleneck.
This is especially true when attempting to capture all relevant real-world situations that the AVs could encounter, such as degraded conditions \cite{zhang2023perception}.
To reduce dependence on high-quality labels, researchers have proposed approaches such as pseudolabels \cite{caine2021pseudolabeling,houston2021one} or unsupervised methods \cite{zhang_cvpr2023_oyster}, which use limited data to help bootstrap complex detection systems.
While useful, these approaches largely supplement, not replace, data-hungry methods.
A more direct line of research, which is also the focus of our work, is improving auto-labels \cite{qi2021offboard}, which are algorithmic labels computed offline that serve as seeds for manual labeling, or even as direct training labels if some loss of quality is acceptable.

VLMs are becoming increasingly popular in the self-driving industry, mirroring their overall rise in popularity.
Researchers have explored VLMs for full end-to-end driving \cite{chen2024wayve}, and to improve driver models via a student-teacher setup for robustness \cite{xu2024vlm}.
While most published work focuses on onboard VLM applications, in this paper we investigate their less-explored offboard use to improve auto-labeling. 
Earlier efforts include \cite{openannotate3d} and \cite{openbox} where authors proposed complex two-stage processes to convert 2D VLM outputs into 3D bounding boxes, unlike our work which directly outputs 3D dimensions.

Another relevant research area is vehicle make and model recognition (VMMR).
This is a perception task that goes beyond just detecting vehicles, by also inferring a vehicle's make and model (e.g., "Toyota RAV4", "Ford Fusion") \cite{tafazzoli2017large}.
VMMR has a long history of research \cite{Gayen2024TwoDecades}, and is an important part of applications such as traffic surveillance and monitoring \cite{manzoor2019real}.
VMMR approaches have transitioned from traditional ML and computer vision methods \cite{pearce2011automatic,siddiqui2016real} to state-of-the-art deep learning methods \cite{lee2019real, tafazzoli2017large}.
A recent survey paper \cite{miu2025can} discussed using VLMs for VMMR, albeit without providing implementation or analysis. 
In contrast, we implement a VLM-based solution to the VMMR problem and are the first to use its outputs to improve auto-labeling.

\section{Methodology}
We assume there exists a system for 3D labeling of traffic actors that provides human labelers with multi-sensor inputs (including multi-camera images and lidar returns), along with seed 3D bounding boxes (usually created by an offboard detector \cite{qi2021offboard}).
Labelers use this system to visualize and navigate traffic scenes, and to modify the bounding boxes so they tightly fit the actors at each timestep.
Our task is to improve and speed up this process in the case of vehicle labels.
In particular, we propose to improve the accuracy of seed 3D bounding boxes by considering vehicle factory dimensions, thus reducing the manual effort needed to produce high-quality labels.

\begin{figure}[t]
    \centering
    \includegraphics[ width=0.98\linewidth]{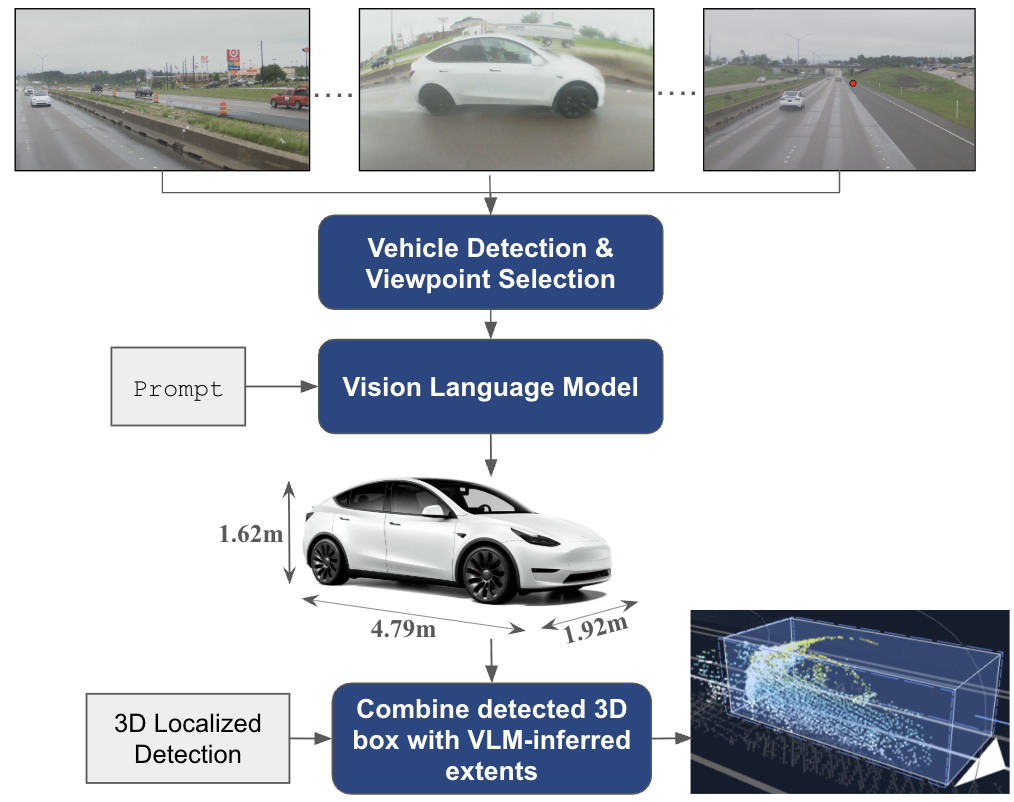}
    \caption{Overview of the proposed auto-labeling framework; examples of image crops resulting from viewpoint selection are shown in Figure \ref{fig:VMMGR_examples}.}
    \label{fig:methodology}
\end{figure}

For the offboard detector we assume an off-the-shelf vehicle detection system relying on available sensors, outputting localized 3D detections and approximate 2D bounding boxes from which we can derive camera crops.
These 2D image crops of individual vehicles are then fed to a pretrained VLM.
The VLM is prompted to perform the coupled task of vehicle make, model, and generation recognition (henceforth referred to as VMMGR) and to output the corresponding factory dimensions. 
The additional generation output is critical, as dimensions and visual design are identical or near-identical within a generational year range (e.g., ``Toyota RAV4 4th generation 2013-2018") but can significantly differ between generations even for the same make and model \cite{CarmaxGen, CargurusGen}. 
The output dimensions are then combined with localized 3D detections and used as seed values for the 3D bounding boxes in the aforementioned labeling system.
Figure \ref{fig:methodology} illustrates this process, from the initial 2D camera inputs to the final, VLM-seeded 3D bounding box.

\subsection{Image Sampling}
Following the initial detection step, we obtain multiple frames of a vehicle across both time and various camera viewpoints. 
Given the limited context window of current state-of-the-art VLMs, we select a subset of frames as input. 
Our sampling strategy is a two-step process. First, we perform viewpoint selection: at each timestamp, we identify a camera that provides the best view of the vehicle, determined by selecting the camera view where the vehicle's projected detection area (in pixels) is largest. 
Second, we uniformly sample $N$ images from these best-view frames over the vehicle's observed time duration, where $N$ is the maximum input capacity of a considered VLM.

\subsection{VLM Setup and Prompt Engineering}
\label{sect:methodology_prompts}
As shown in Figure \ref{fig:methodology}, a sequence of image crops is provided to the VLM to determine vehicle dimensions. 
We performed several iterations of prompt engineering to increase dimension accuracy; our prompts evolved from requesting only dimensions to also including vehicle make, model, and generation recognition (VMMGR).
Each iteration involved reviewing dimension and VMMGR performance metrics, analyzing failures, and refining the prompt. 
Here we list important iterations, with detailed results discussed in Section \ref{sect:analysis_prompt_engineering}:
\begin{enumerate}
    \item \textbf{Basic Prompt}: A short query prompt that directly asks the VLM for vehicle dimensions given the image sequence, with no additional guidance.
    \item \textbf{Vehicle Type Prompt}: The prompt asks the VLM to first classify the vehicle into high-level types (sedan, hatchback, SUV, pickup, van) and then use this classification to predict vehicle dimensions. 
    \item \textbf{Type + Market Size Class Prompt}: In addition to high-level type, this prompt asks the VLM to also determine market size class (e.g., subcompact, compact, midsize, full-size) and use both for dimension prediction. 
    \item \textbf{VMMGR Prompt}: Prompts the VLM to reason about vehicle type, market size class, and, critically, the vehicle’s make, model, and generation year range. 
    \item \textbf{Refined VMMGR Prompt}: We added several other reasoning elements into the VMMGR prompt, inspired by the chain-of-thought engineering \cite{chainofthought}. This final prompt incorporated the following steps:
        \begin{enumerate}
            \item Occlusion assessment: To prevent failures from significant occlusion (which we found led to hallucinated outputs or predictions on an occluding vehicle), we added a step for the VLM to first assess if it can confidently identify the vehicle due to potential occlusion.
            \item Make, model, generation identification: This is the core VMMGR task from the previous prompt, asking for the vehicle's make, model, and generation.
            \item Vehicle configuration identification: Because we found that dimensions were less accurate for vehicles with multiple configurations (e.g., pickup bed length, van wheelbase), we requested an additional output to identify these specific configurations.
            \item Modification and damage assessment: To handle vehicles differing from factory specifications (found to be common poor performers), the prompt asks the VLM to identify any visible modifications or damage (e.g., lifted suspension, wider fenders) and to state whether each dimension (i.e., length, width, height) is increased, decreased, or fits factory specifications.
            \item Factory dimension retrieval: Finally, the prompt requests the VLM to output the vehicle's factory dimensions (length, width, height) in meters, based on its preceding reasoning.
        \end{enumerate}     
\end{enumerate}

For all prompts, we ask the VLM to output the fields specified in a JSON template included within the prompt, which can be set to null if unsure. 
Specifically, the complete final refined VMMGR prompt is as follows:

{\tiny
\begin{verbatim}
## SYSTEM PROMPT
You are an expert vehicle perception system for an autonomous vehicle. Your primary
function is to analyze sequences of image frames to identify vehicles and their
attributes with high accuracy.

Your analysis must be robust. The input will be multiple frames from a video where
quality, angle, and distance may vary. You must base your conclusions on the best
available evidence across all provided frames, first determining if a clear
identification is even possible due to occlusion.

You will identify the vehicle's make, model, configuration, and generation,
retrieve its standard factory dimensions, and critically assess it for any
modifications or damage that would alter those dimensions. Your output must
strictly adhere to the requested JSON format.

## USER PROMPT
Follow these steps carefully to analyze the vehicle in the provided image frames:

**1. Occlusion Assessment:**
- First, examine all provided frames to determine if the target vehicle is
identifiable.
- Set **`significantly_occluded`** to `true` if, and only if, the vehicle is so
severely blocked by other cars or objects in **every single frame** that its make,
model, and shape cannot be confidently determined.
- If `significantly_occluded` is `true`, fill all other fields in the JSON with
`null` and stop.
- If at least **one frame** offers a clear enough view for identification, set
`significantly_occluded` to `false` and proceed to the next step.

**2. Frame Synthesis & Vehicle Identification:**
a. **Primary ID:** Synthesize information from the clearest frames to identify the
vehicle's **make**, **model**, and **generation** (e.g., "2019-2024").
b. **Configuration ID:** For vehicles like pickup trucks, full-size SUVs, or vans,
identify the specific **configuration**.
This could be cab size (Crew Cab, Double Cab), bed length (Short, Standard, Long),
or wheelbase (e.g., Tahoe vs. Suburban).
**Prioritize side-view frames** to make this determination.

**3. Factory Dimension Retrieval:**
- Based on the identified generation AND configuration, provide the standard
**factory** dimensions (length, width, and height) in meters.

**4. Modification & Damage Assessment:**
- Inspect the vehicle for visible non-factory modifications or significant damage.
- **Height changes:** Lift kits, suspension lowering.
- **Width changes:** Wide-body kits, aftermarket fender flares, wider tires.
- **Length changes:** Aftermarket bumpers, externally mounted spare tires, bike racks.
- **Important:** A standard tow hitch should **not** be flagged as a length
modification.

**5. JSON Output Generation:**
- Format your final analysis as a single JSON object using the schema below.
- If the configuration cannot be confidently determined, set the `configuration` value
to "Best estimate - configuration unclear", but still provide your best estimate for
the dimensions.
- For modification keys, use "increased" or "decreased" only if you observe a relevant 
modification; otherwise, use null.
}

json
{
    "significantly_occluded": "boolean",
    "make": "string or null",
    "model": "string or null",
    "generation_year_range": "string or null",
    "vehicle_type": "string ('sedan', 'hatchback', 'SUV', 'pickup truck', 'van',
    'other') or null",
    "configuration": "string or null",
    "length_m": "float or null",
    "width_m": "float or null",
    "height_m": "float or null",
    "length_modification": "'increased', 'decreased', or null",
    "width_modification": "'increased', 'decreased', or null",
    "height_modification": "'increased', 'decreased', or null"
}
\end{verbatim}
}

\begin{table*}[!t]
    \centering
    \caption{Performance comparison of VLMs using the Refined VMMGR prompt \\(lower is better for error metrics, higher is better for IoU and percentage predictions; best results shown in bold)}
    \vspace{-0.20cm}
    
    \begin{tabular*}{\textwidth}{@{\extracolsep{\fill}}l ccccccccc}
        \toprule
        \textbf{Metric} & 
        \multicolumn{1}{c}{\textbf{Baseline}} & \multicolumn{1}{c}{\textbf{Llama 4 Maverick}} & \multicolumn{1}{c}{\textbf{Pixtral Large}} & \multicolumn{2}{c}{\textbf{Claude Sonnet 4}} & \multicolumn{2}{c}{\textbf{Gemini Flash 2.5}} & \multicolumn{2}{c}{\textbf{Gemini Pro 2.5}}  \\
        
        \midrule

        \bm{$N$} (\# of context images) & \multicolumn{1}{c}{\textbf{-}} &
        \multicolumn{1}{c}{\textbf{3}} & \multicolumn{1}{c}{\textbf{3}} &
        \multicolumn{1}{c}{\textbf{3}} & \multicolumn{1}{c}{\textbf{10}} & \multicolumn{1}{c}{\textbf{3}} &
        \multicolumn{1}{c}{\textbf{10}} & \multicolumn{1}{c}{\textbf{3}} &
        \multicolumn{1}{c}{\textbf{10}}\\

        \midrule
        
        Absolute length error (meters) & 0.3010 & 0.2946	 & 0.2837 & 0.3142 & 0.2911		 & 0.2678 & 0.2610 & 0.2635 & \textbf{0.2589} \\
        Absolute width error (meters) & 0.1157 & 0.0957	 & 0.0909 & 0.0885 & 0.0853		& 0.0805	 & 0.0786	 & 0.0778	 & \textbf{0.0767}	  \\
        Absolute height error (meters) & 0.0939 & 0.1025 & 0.0950 & 0.0995 & 0.0933		& 0.0784 & 0.0764	 & 0.0765	& \textbf{0.0752}	 \\
        Relative length error (proportion) & 0.0634 & 0.0622  & 0.0602& 0.0668 & 0.0619	 & 0.0567	& 0.0554	 & 0.0560 & \textbf{0.0549}\\
        Relative width error (proportion) & 0.0557 & 0.0462	& 0.0438  & 0.0427 & 0.0411 & 0.0388 & 0.0379	 & 0.0376 & \textbf{0.0370}	\\
        Relative height error (proportion) & 0.0494 & 0.0556  & 0.0515	& 0.0537 & 0.0501		& 0.0422 & 0.0412 & 0.0412 & \textbf{0.0405}	 \\
        Intersection-over-union & 0.8497 & 0.8559	 & 0.8635 & 0.8581 & 0.8663		 & 0.8794	& 0.8813		 & 0.8808 & \textbf{0.8831}  \\
        Percentage predictions & \textbf{100\%} & 86.92\% & 96.91\% & 93.10\% & 97.36\% & 99.03\% & 99.63\% & 98.82\% & 99.26\% \\
        \bottomrule
    \end{tabular*}
    
    \label{tab:different_llms}
    \vspace{-0.4cm}
\end{table*}

\begin{table*}[!t]
    \centering
    
    \caption{Comparison of VLM accuracy on the VMMGR task, sliced by vehicle type (higher is better, best results shown in bold) 
    }
    \vspace{-0.15cm}
    
    \begin{tabular*}{\textwidth}{@{\extracolsep{\fill}}l ccccc c}
        \toprule
        \textbf{Model} & \multicolumn{1}{c}{\textbf{Sedan}} & \multicolumn{1}{c}{\textbf{SUV}} & \multicolumn{1}{c}{\textbf{Pickup Truck}} & \multicolumn{1}{c}{\textbf{Van}} & \multicolumn{1}{c}{\textbf{Hatchback}} & \multicolumn{1}{c}{\textbf{Total}} \\
        \midrule
        
        Llama 4 Maverick & 70.5\% & 65.3\% & 81.3\% & 80.6\% & 88.0\% & 74.5\% \\
        Pixtral Large & 72.4\% & 65.8\% & 78.0\% & 85.6\% & 68.0\% & 74.2\% \\
        Claude Sonnet 4 & 89.7\% & 79.3\% & 87.9\% & 98.0\% & \textbf{100.0\%} & 88.4\% \\
        Gemini Flash 2.5 & 97.3\% & 94.0\% & 97.3\% & \textbf{99.0\%} & 96.0\% & 96.7\% \\
        Gemini Pro 2.5 & \textbf{100.0\%} & \textbf{95.3\%} & \textbf{98.7\%} & \textbf{99.0\%} & 96.0\% & \textbf{98.1\%} \\
        
        \bottomrule
    \end{tabular*}
    
    \label{tab:VMMGR_metrics}
    \vspace{-0.4cm}
\end{table*}

\section{Experiments}

\subsection{Experimental Setup}
We used two sources of 3D labeled data to test the proposed approach, including an internal proprietary dataset and an external open-source dataset.
The internal dataset consists of 15-second-long snippets collected on US highways and surface streets in Texas, Pennsylvania, and California between 2022 and 2025. 
The snippets comprise camera, lidar, and radar sensor data, and were labeled with 3D bounding boxes and vehicle types by an internal team of professional labelers.
The bounding boxes are drawn in a way to encompass the entire vehicle body, excluding minor protrusions (e.g., small side mirrors).
We randomly sampled $3{,}821$ vehicle labels from these snippets for our experiments.
In addition, we used the Waymo Open Dataset \cite{sun2020waymo} as an external open-source dataset.
To ensure high label quality we considered only actors within $55m$ of the AV, resulting in $1{,}931$ vehicle samples.

For VLMs, we evaluated several state-of-the-art pretrained VLMs available through publicly accessible, third-party cloud services, without any model finetuning.
In particular, we used Llama 4 Maverick \cite{llama4}, Pixtral Large \cite{MistralAIPixtralLarge}, Claude Sonnet 4 \cite{claude4}, as well as Gemini Flash 2.5 and Pro 2.5 \cite{gemini25}.

For fair comparison on bounding box inference, we designed a strong non-VLM oracle baseline. Critically, this baseline uses the label's ground-truth vehicle type during inference to predict fixed dimensions for each of the five most common types of vehicles (covering more than $99\%$ of all vehicles), namely sedan, SUV, pickup truck, van, and hatchback. These fixed dimensions were determined as follows.
Each vehicle type has standard size classes (e.g., for sedans: subcompact, compact, midsize, full-size).
Most manufacturers' vehicle dimensions in each size class are fairly similar, typically down to a few centimeters \cite{JDPower2024Quality,CarAndDriver2025SUVs}. For each vehicle type and size class, we retrieve the dimensions from the best-selling vehicles in that class. To account for the model year, we used the dimensions of 2013 models, approximating the average vehicle age in the US as of 2025 \cite{SPGlobal2025Age}. 
The final predicted dimension for a type is the mean of the size class dimensions.
For the $1\%$ of vehicles that do not fall into one of the five main types (e.g., convertibles, RVs, trailers), we output a mean bounding box computed over all predictions made by the baseline on the five main types.

\subsection{Analysis of Different VLM Approaches}
We first compared the performance of different VLMs on the internal data, using the refined VMMGR prompt which yielded the best results in all cases.
In addition, we used a maximum allowed number of context images $N$ for each VLM, which was $3$ for Llama and Pixtral and $10$ for Claude and Gemini. 
For a direct comparison, we also conducted experiments with $3$ images for Claude and Gemini.

\subsubsection{Inferring Bounding Box Size}
Since the prompt allows the VLM to abstain from prediction (e.g., due to occlusion), we report the "Percentage predictions" metric, which is the percentage of samples where it returned non-null dimensions. 
For non-predictions, we imputed the mean of the model's predicted dimensions for all other samples. 
The results are summarized in Table \ref{tab:different_llms}, where we report the average values of absolute and relative errors along the three dimensions (length, width, height), as well as the intersection-over-union (IoU) metric computed in the 2D overhead view.

We can see that the baseline shows very competitive performance, achieving low error rates across all metrics. 
Unlike the VLM methods which can abstain from outputting the dimensions, it also achieves a $100\%$ prediction rate. 
However, it is crucial to note that the baseline acts as an oracle, using the ground-truth actor type to infer the vehicle size.

Focusing first on the $N=3$ results, the VLM methods in general improved over this strong baseline. 
For example, Llama, Pixtral, and Claude mostly improved on the length and width metrics compared to the baseline, while all showed worse performance on height. 
The Gemini 2.5 models clearly outperformed both the baseline and the other VLMs, with Gemini Pro achieving the strongest performance.
In addition, the $N=10$ results for Claude and Gemini show a clear benefit from the longer context, improving all considered metrics when compared to the shorter-context variants.
In particular, Gemini Pro with $N=10$ achieved the lowest errors across all metrics, the highest IoU, and a near-perfect percentage prediction.
Given these results, in the following experiments we set $N$ to a maximum context allowed by a particular VLM.

\begin{figure*}[!t]
    \centering 

    \subfloat[This 1st-generation Chrysler 300 sedan was $88m$ away at its closest, without visible logo or branding, yet correctly identified by all models.]{\includegraphics[width=0.23\textwidth]{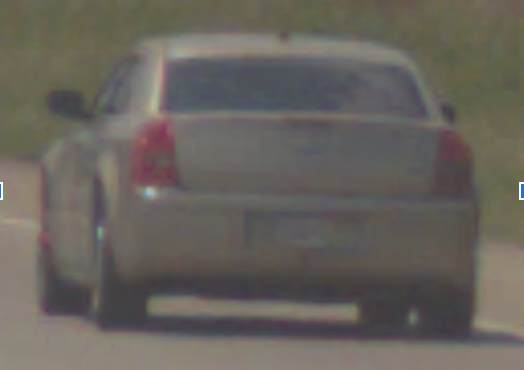}\label{fig:sub1}}
    \hfill 
    \subfloat[Ford Focus came in two versions in the US: a sedan and a hatchback; all VLMs correctly identified it as the 3rd-generation hatchback.]{\includegraphics[width=0.23\textwidth]{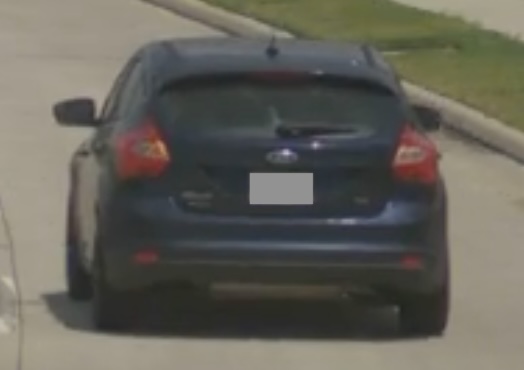}\label{fig:sub2}}
    \hfill
    \subfloat[Ford Transit has two configurations that differ in length by $2m$; all VLMs identified this $90m$-away vehicle as the shorter Transit Connect.]{\includegraphics[width=0.23\textwidth]{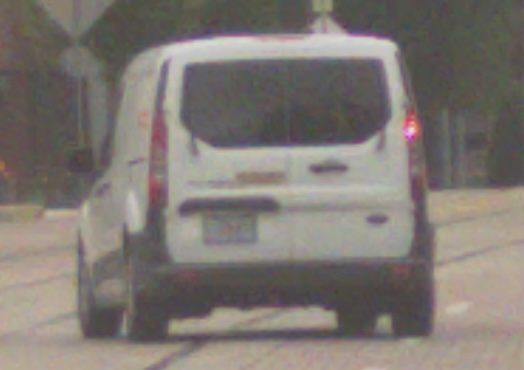}\label{fig:sub3}}
    \hfill
    \subfloat[All models correctly identified the 30-year-old vehicle as Chevrolet Silverado 1990-1998 from the front view alone.]{\includegraphics[width=0.23\textwidth]{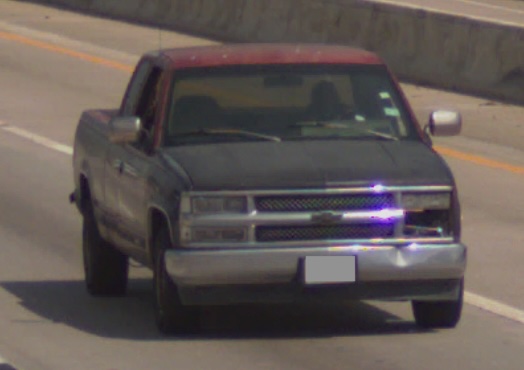}\label{fig:sub4}}

    \vspace{0.25em} 

    \subfloat[Nissan sedans of this year range have very similar front styling, making identification challenging; Llama incorrectly inferred it as Sentra, while the other VLMs correctly identified a 5th-generation Altima.]{\includegraphics[width=0.23\textwidth]{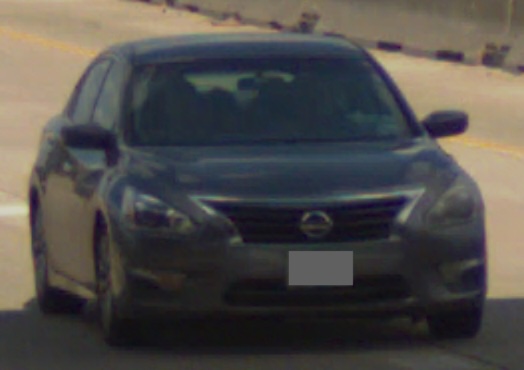}\label{fig:sub5}}
    \hfill
    \subfloat[This sedan is 3 lanes away, having very generic styling; Llama, Pixtral, and Claude incorrectly predicted common Toyota and Hyundai sedans, while Gemini models correctly identified a 2nd-gen Kia Forte.]{\includegraphics[width=0.23\textwidth]{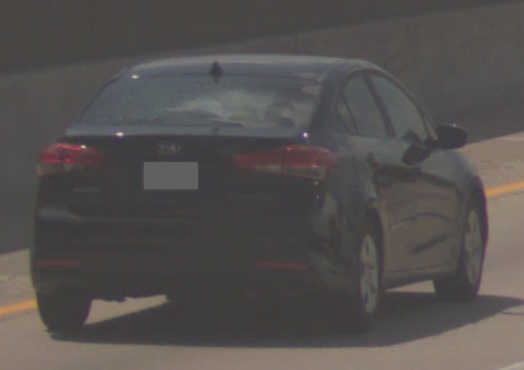}\label{fig:sub6}}
    \hfill
    \subfloat[This Hyundai Santa Fe 2018-2022 is $90m$ away at its closest point; all VLMs output wrong prediction (Llama: Volvo XC60, Pixtral: Toyota RAV4, Claude: Honda CR-V, Gemini: Buick Envision).]{\includegraphics[width=0.23\textwidth]{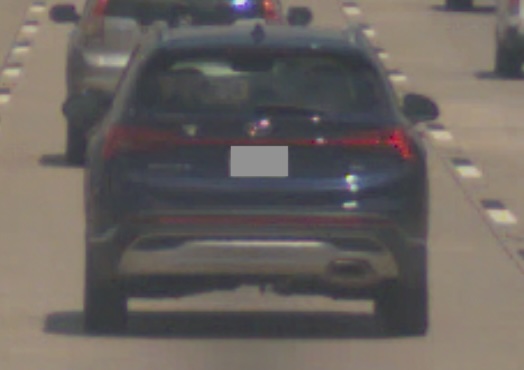}\label{fig:sub7}}
    \hfill
    \subfloat[This vehicle is partially occluded, with generic styling: all VLMs predicted different Toyota SUVs, but it is actually a 2nd-generation Toyota Sienna minivan.]{\includegraphics[width=0.23\textwidth]{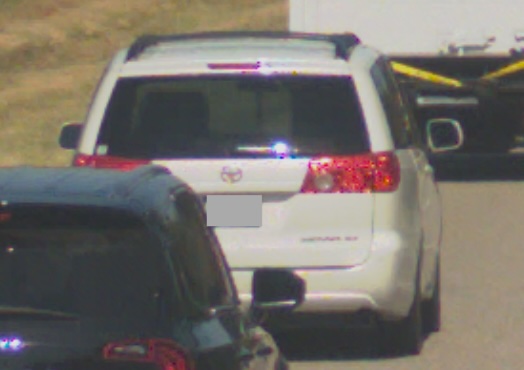}\label{fig:sub8}}
    
    \caption{Examples of VMMGR predictions, with each image showing the clearest camera crop from the input image sequence; the top row shows cases where all VLMs provided correct predictions, while the bottom row shows interesting failure modes.}
    \label{fig:VMMGR_examples}

    \vspace{-0.4cm}
\end{figure*}

\subsubsection{Inferring Vehicle Make/Model/Generation}
In Table \ref{tab:VMMGR_metrics} we explore the VLM performance on the VMMGR task, sliced by the vehicle type. A VMMGR prediction was counted as correct only if the make, model, and generation were all correct.
Consistent with the previous experiment, Gemini Pro 2.5 achieved the best overall performance, with the exception of hatchbacks (although still reaching competitive accuracy).
Performance varied significantly across vehicle types, with hatchbacks and vans being the easiest to infer correctly for all VLMs, while SUVs and sedans presented the biggest challenge. 
Such results can be explained by the level of distinguishing visual features that each type has, as explored further in Figure \ref{fig:VMMGR_examples} which shows some notable examples where the VLMs returned correct as well as incorrect predictions.

\begin{figure}[!t]
    \centering 

    \subfloat[Wider and taller vehicle due to oversized tires and lifted suspension.]{\includegraphics[width=0.48\columnwidth]{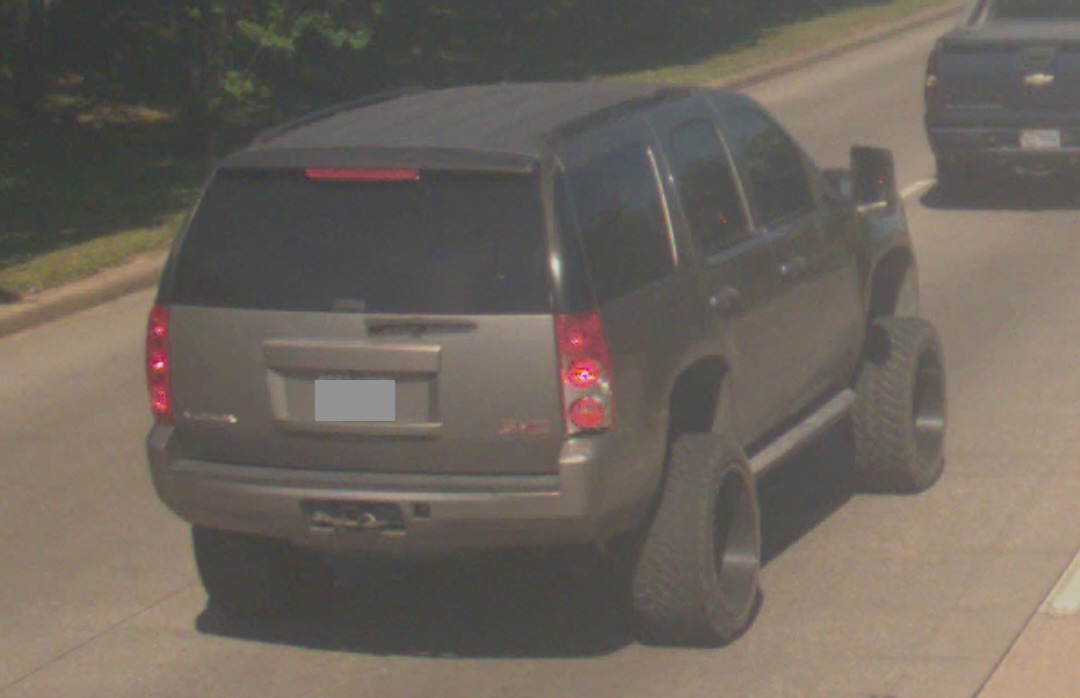}\label{fig:mod1}}
    \hfill 
    \subfloat[Larger in all three dimensions due to protruding tires and cargo racks.]{\includegraphics[width=0.48\columnwidth]{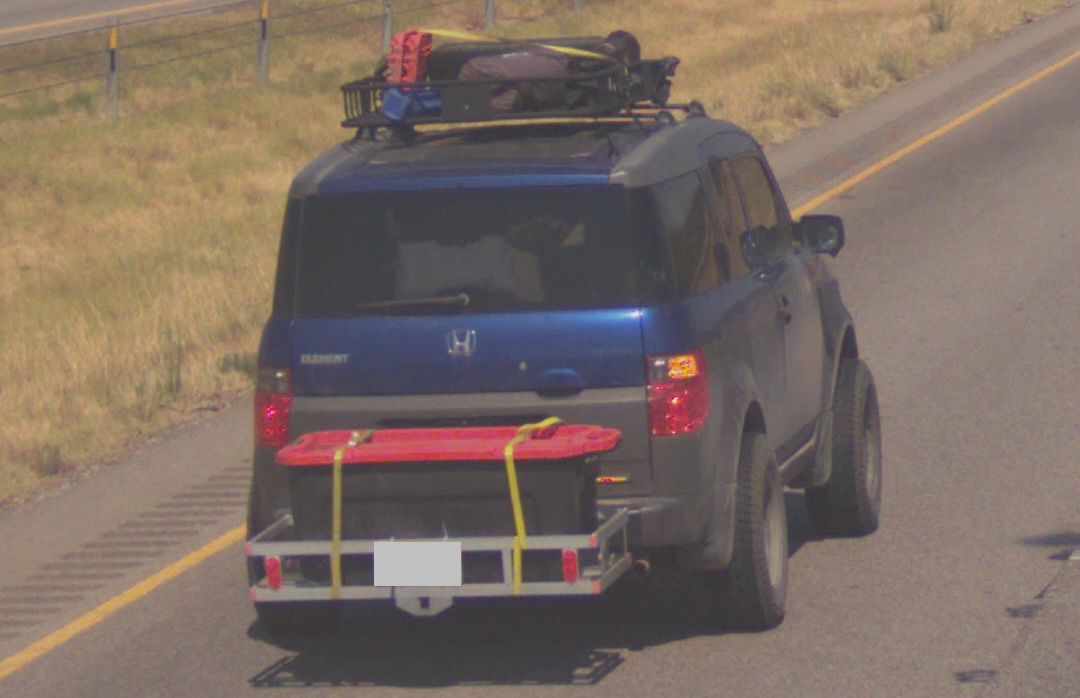}\label{fig:mod2}}


    
    \caption{Examples of SUVs identified as being modified.}
    \label{fig:modification_examples} 
    \vspace{-0.4cm}
    
\end{figure}

\subsubsection{Modification Prediction}

An important output of the system is the identification of modifications or damage that could alter the factory dimensions. 
This allows the system to identify samples that likely may not match their factory specs.

We evaluated the modification output using Gemini Pro 2.5 which identified $82.93\%$ of actors as unmodified, with the remaining $17.07\%$ as having some modification.
A comparison of absolute error metrics confirmed the importance of this feature: the unmodified vehicles had significantly lower errors ($0.250m$/$0.068m$/$0.062m$ in length/width/height), compared to the modified actors ($0.302m$/$0.119m$/$0.140m$).
Illustrative cases of vehicles found to be modified are shown in Figure \ref{fig:modification_examples}.


In the context of auto-labeling, these findings show that the modification prediction has promising potential to be used for:
\begin{itemize}
    \item Intelligently flagging or filtering out predictions where the system believes it may be less accurate, resulting in significantly more accurate auto-labels;
    \item Surfacing interesting outliers that are modified or damaged, allowing for special focus on these long-tail vehicles during labeling and model training.
\end{itemize}

\begin{table*}[!t]
    \centering
    
    \caption{Performance comparison of baseline vs. different prompts using Gemini Pro 2.5 (best results shown in bold)}
    
\begin{tabular*}{\textwidth}{@{\extracolsep{\fill}}l cccccc}
        \toprule
        \textbf{Metric} & \multicolumn{1}{c}{\textbf{Baseline}} & \multicolumn{1}{c}{\textbf{Basic}} & \multicolumn{1}{c}{\textbf{Vehicle Type}} & \multicolumn{1}{c}{\textbf{Type + Size Class}} & \multicolumn{1}{c}{\textbf{VMMGR}} & \multicolumn{1}{c}{\textbf{Refined VMMGR}} \\
        \midrule
        
        Absolute length error (meters) & 0.3010& 0.3059 & 0.3128 & 0.2808 & 0.2647 & \textbf{0.2589}	\\
        Absolute width error (meters) & 0.1157 & 0.0772 & 0.0991 & 0.0770 & \textbf{0.0767} & \textbf{0.0767} \\
        Absolute height error (meters) & 0.0939 & \textbf{0.0699} & 0.0996 & 0.0826 & 0.0755 & 0.0752 \\
        Relative length error (proportion) & 0.0634 & 0.0638 & 0.0665 & 0.0597 & 0.0561 & \textbf{0.0549} \\
        Relative width error (proportion) & 0.0557 & 0.0375 & 0.0478 & 0.0371 & 0.0371 & \textbf{0.0370} \\
        Relative height error (proportion) & 0.0494 & \textbf{0.0382} & 0.0511 & 0.0447 & 0.0530 & 0.0405 \\
        Intersection-over-union & 0.8497 & 0.8827 & 0.8550 & 0.8766 & 0.8830 & \textbf{0.8831} \\
        Percentage predictions & \textbf{100\%} & 97.85\% & 99.89\% & 98.06\% & 99.86\% & 99.26\% \\
        \bottomrule
    \end{tabular*}
    
    \label{tab:prompt_engineering2}
\end{table*}

  
    
    
  

\begin{table}[h!] 
  \centering
  
  \caption{Performance on the open-sourced Waymo dataset}
  \label{tab:waymo_data}
  
  \begin{tabular}{l c}
    \toprule
    \textbf{Metric} & \textbf{Gemini Pro 2.5} \\ 
    \midrule

        Absolute length error (meters) & 0.2654	\\ 
        Absolute width error (meters) & 0.2475	\\ 
        Absolute height error (meters) & 0.1396 \\ 
        Relative length error (proportion) & 0.0678 \\ 
        Relative width error (proportion) & 0.1216 \\ 
        Relative height error (proportion) & 0.0709 \\ 
        Intersection-over-union & 0.7715 \\ 
        Percentage predictions & 91.21\% \\ 

    \bottomrule
  \end{tabular}

\end{table}

\subsection{Analysis of Prompt Engineering}
\label{sect:analysis_prompt_engineering}
This section provides a detailed ablation study of the various prompts introduced in Section \ref{sect:methodology_prompts}, with results summarized in Table \ref{tab:prompt_engineering2}. For brevity, we report ablation results for Gemini Pro 2.5, which achieved the strongest overall performance. 
We confirmed, however, that this pattern of improvement was general and applied to all other VLMs used in Table \ref{tab:different_llms}.

The basic prompt, which directly asks for vehicle dimensions, already achieved reasonable performance, although not always beating the baseline (such as in the length metrics). 
Interestingly, the vehicle type prompt, which asks the VLM to use the vehicle type to infer dimensions, actually led to reduced performance. 
This may be due to the prompt artificially limiting the VLM's reasoning capabilities by explicitly linking the dimensions to only the high-level vehicle type.

The next iteration, which added reasoning about the vehicle size class (e.g., subcompact, compact), significantly improved on the previous prompt and outperformed the basic prompt on all metrics except height. 
This is not surprising as, for instance, SUVs can significantly differ in size, with their length being under 4 meters for a subcompact SUV to nearly 6 meters for a full-size SUV, which is a variance that reasoning about size class can capture.
The introduction of explicit VMMGR reasoning further improved nearly all metrics compared to the ``Type + Size Class" prompt, demonstrating the benefits of this added granularity.
Lastly, using even more specific reasoning in the refined VMMGR prompt (such as occlusion reasoning, vehicle configuration, and modifications) yielded another performance boost, resulting in the best overall prompt.

\subsection{Surfacing and Correcting Human Labeling Errors}



\begin{figure}[t]
    \centering
    \subfloat[Significantly-occluded SUV, resulting in lidar coverage on only its rear half; by overlaying an image of the identified 2nd-generation Volvo XC90, we can see that the label could be improved through lengthening by $0.62m$. \label{fig:volvo_dim}]{%
        \resizebox{\columnwidth}{!}{%
            \includegraphics[height=3cm]{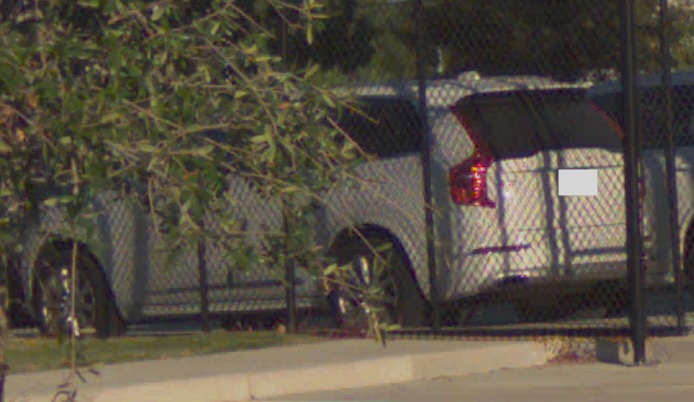}%
            \hspace{0.3cm}%
            \includegraphics[height=3cm]{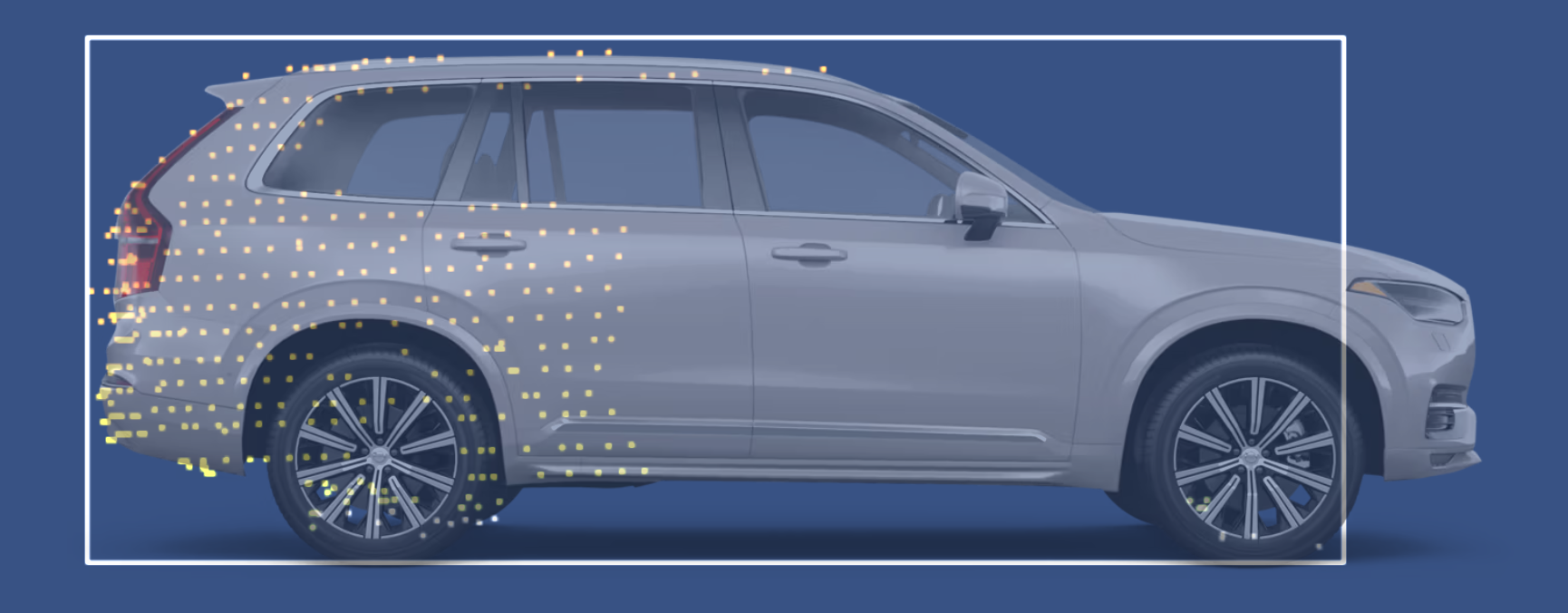}%
        }%
    }


    \subfloat[Commonly found occlusion on front vehicles, where only the rear three-quarters are visible; overlaying the identified Cadillac XTS sedan shows a $0.35m$ portion that was occluded in lidar and thus not included in the label. \label{fig:cadillac_dim}]{%
        \resizebox{\columnwidth}{!}{%
            \includegraphics[height=3cm]{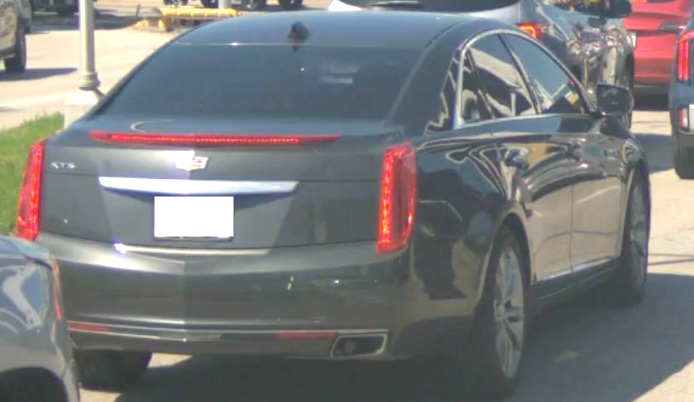}%
            \hspace{0.3cm}%
            \includegraphics[height=3cm]{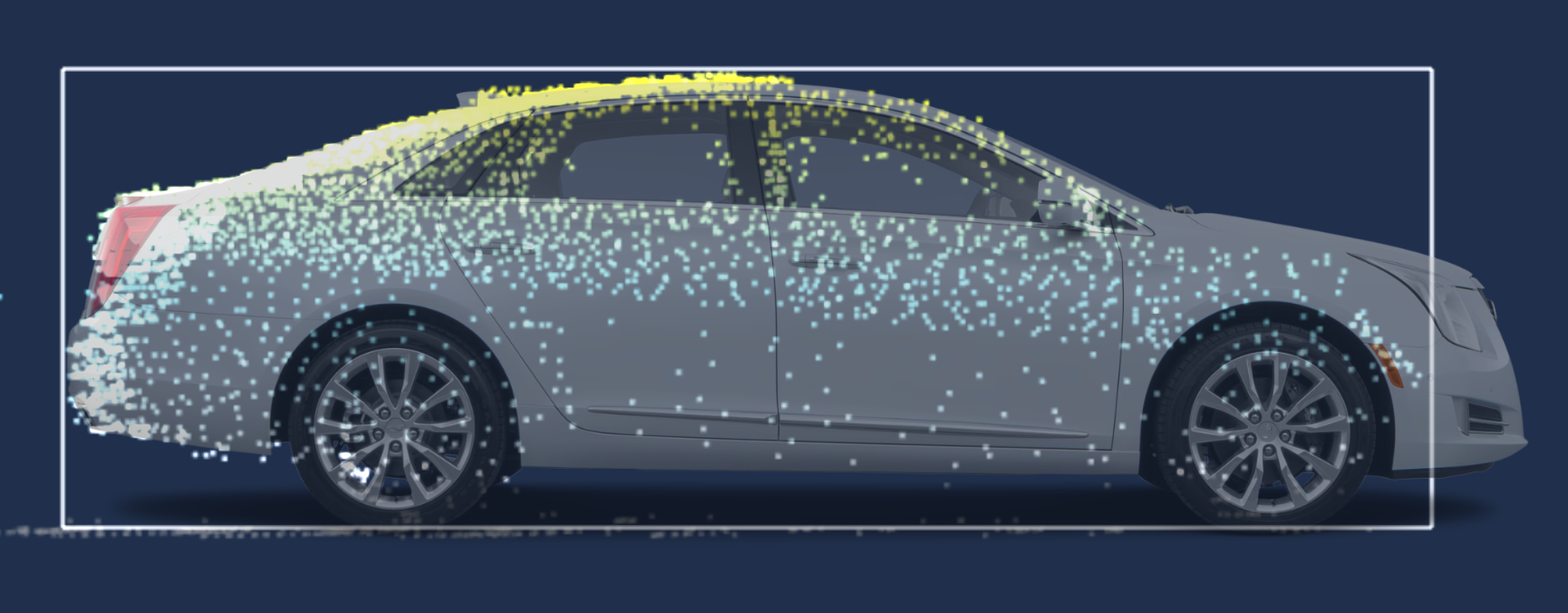}%
        }%
    }

    \caption{Examples where the proposed system identified potential label improvements. Each example shows (left) the input camera view, and (right) a side-view diagram with projected lidar points and label box in white, along with a reference image of the predicted vehicle overlaid to align with lidar.}
    \label{fig:dimension_images}

\end{figure}

We considered the human-provided labels as ground-truth so far, assuming they were perfect (as is a common practice).
However, analysis of high-error predictions revealed that for a small number of these cases, the errors could be explained by suboptimal human labels, as opposed to being due to incorrect predictions.
Figure \ref{fig:dimension_images} highlights notable examples where the proposed approach identified potential label improvements.
We can see that, for both minor and major occlusion cases, the VLM-based method can surpass the initial lidar-aided human-annotated labels and provide very accurate dimensions, highlighting its potential to create high-quality data. 

\begin{figure}[t]
    \centering
    \resizebox{\columnwidth}{!}{%
        \includegraphics[height=3cm]{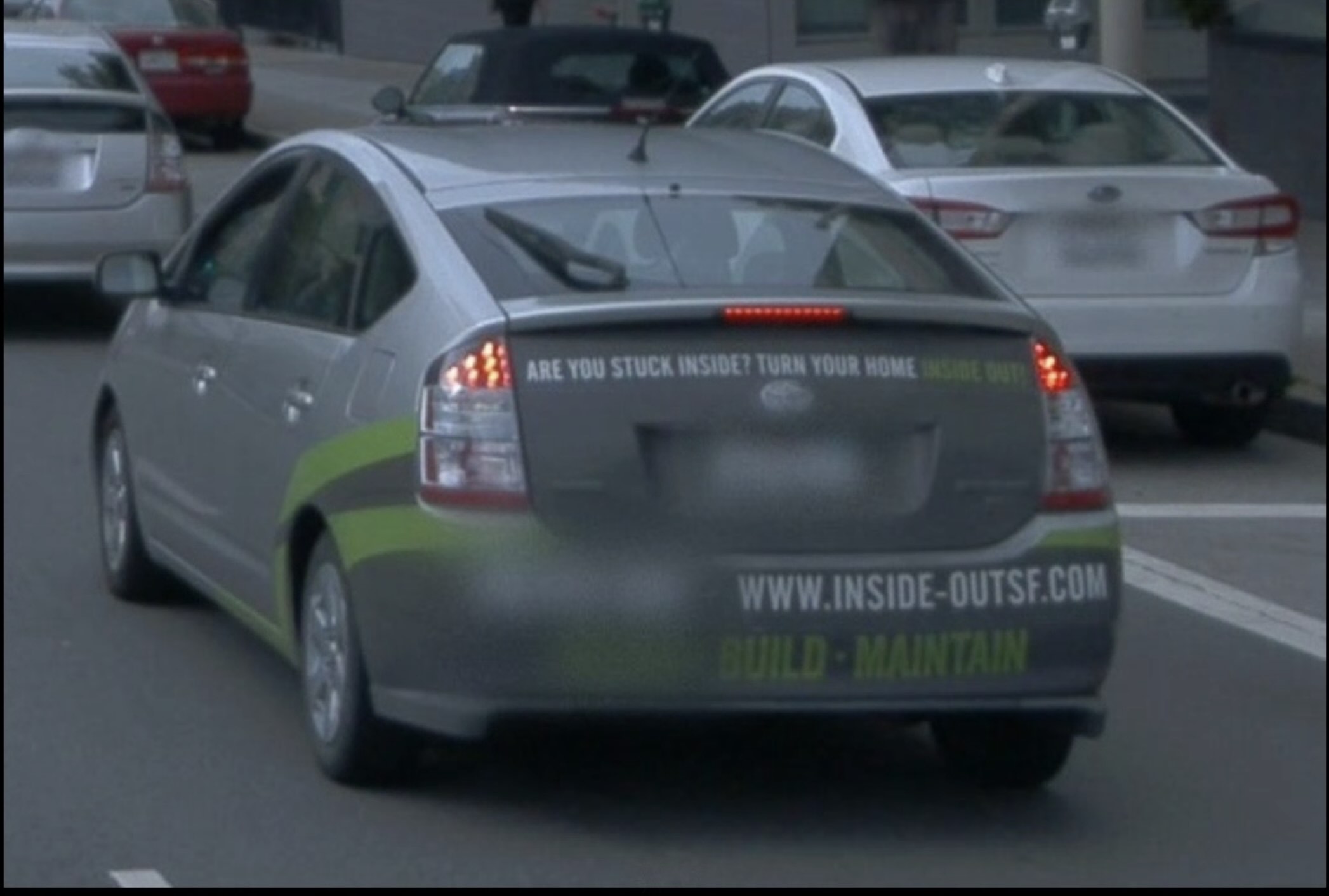}%
        \hspace{0.3cm}
        \includegraphics[height=3cm]{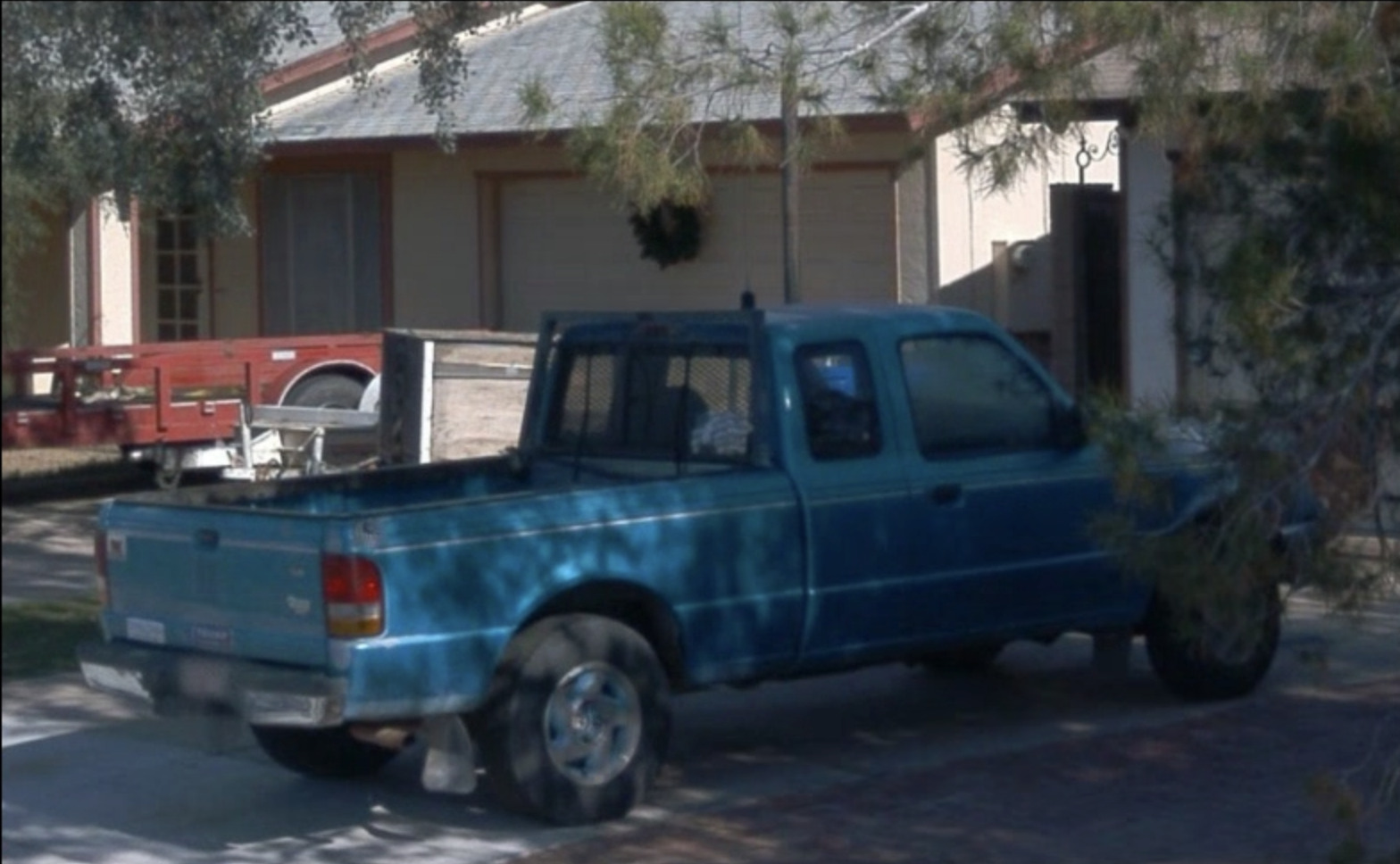}%
    }
    \caption{Examples from the Waymo dataset: (a) Front vehicle correctly identified as a 3rd-generation Toyota Prius; however, the label is $0.4m$ wider than its factory width due to included side-view mirrors; (b) Correctly identified Ford pickup truck has height label of $2.72m$ which is $0.92m$ more than the expected height, seemingly due to the antenna sticking out.}
    \label{fig:waymo_examples}

\end{figure}

\subsection{Experiments on Open-Sourced Data}
Previous experiments were conducted on proprietary data; however, we found that similar conclusions can be derived by evaluating on open-source data.
To that end, we analyzed the results of Gemini Pro 2.5 with the refined VMMGR prompt on the Waymo Open Dataset \cite{sun2020waymo}, summarized in Table \ref{tab:waymo_data}. 

The results for length roughly align with the ones shown in Table \ref{tab:prompt_engineering2}. 
However, we observed a much higher error for width and height. 
Detailed analysis found that in many cases the model predictions were more aligned with the images than the provided labels.
In particular, the predicted width and height are nearly always smaller than the label’s width and height. 
This can be explained by the fact that the model predictions cover factory dimensions, ignoring protrusions such as side-view mirrors and antennas which may be included in the Waymo labels, as illustrated in Figure \ref{fig:waymo_examples}.

These findings suggest that our results are generalizable to other self-driving datasets, and that the proposed approach can more broadly improve the auto-labeling process.

\section{Conclusion}

We proposed a VLM-based approach to improve auto-labeling in self-driving applications. 
Our method uses a single-pass VLM prompt that infers a vehicle's make, model, and generation from camera images to output accurate 3D bounding box dimensions. 
We presented an iterative prompt engineering process and compared our final prompt to a strong oracle baseline and several VLMs, demonstrating the superior performance of the proposed approach. 
Evaluation on both public and proprietary datasets showed the benefits and generalizability of this method. 
Notably, we identified challenging scenarios, such as occlusion, where the VLM-based approach can provide more accurate dimensions than initial lidar-aided human-annotated labels.
The results strongly suggest that integrating VLM-based reasoning into the labeling pipeline can significantly reduce labeling time while increasing label quality.

{
\AtNextBibliography{\footnotesize}
\printbibliography
}

\end{document}